\documentclass[letterpaper]{article}
\usepackage{iccc}
\usepackage{graphicx} 

\usepackage{times}
\usepackage{helvet}
\usepackage{courier}

\usepackage{caption}
\usepackage{subcaption}

\title{AI-Generated Imagery: A New Era for the `Readymade'\\ 
}

\author{
    Amy Smith\textsuperscript{1} and Michael Cook\textsuperscript{2} \\
    \textsuperscript{1}\, School of Electronic Engineering and Computer Science, Queen Mary University of London, UK\\
    \textsuperscript{2}\, Department of Informatics, Kings College London, UK\\
}

\begin{document} 

\maketitle


\begin{abstract}

\begin{quote}
 
 While the term `art' defies any concrete definition, this paper aims to examine how digital images produced by generative AI systems, such as Midjourney, have come to be so regularly referred to as such. The discourse around the classification of AI-generated imagery as art is currently somewhat homogeneous, lacking the more nuanced aspects that would apply to more traditional modes of artistic media production. This paper aims to bring important philosophical considerations to the surface of the discussion around AI-generated imagery in the context of art. We employ existing philosophical frameworks and theories of language to suggest that some AI-generated imagery, by virtue of its visual properties within these frameworks, can be presented as `readymades' for consideration as art.
 
\end{quote}
\end{abstract}

\section{Introduction}

\begin{figure*}[t]
\centering
\begin{subfigure}[b]{0.3\textwidth}
    \centering
    \includegraphics[width=\textwidth]{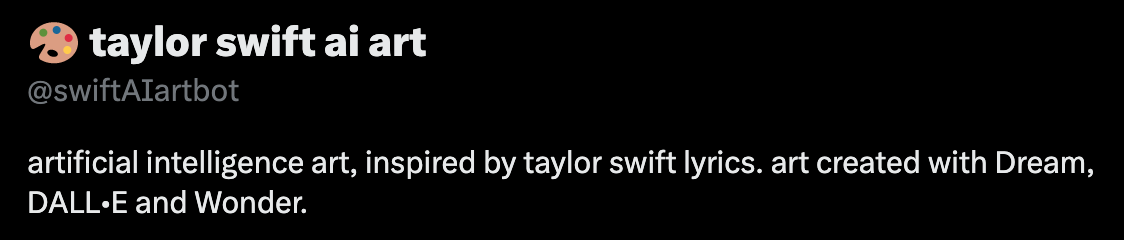}
    \includegraphics[width=\textwidth]{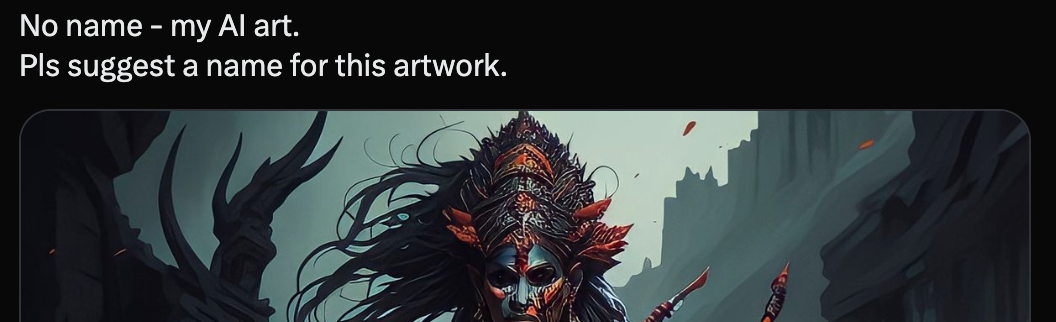}
    \includegraphics[width=\textwidth]{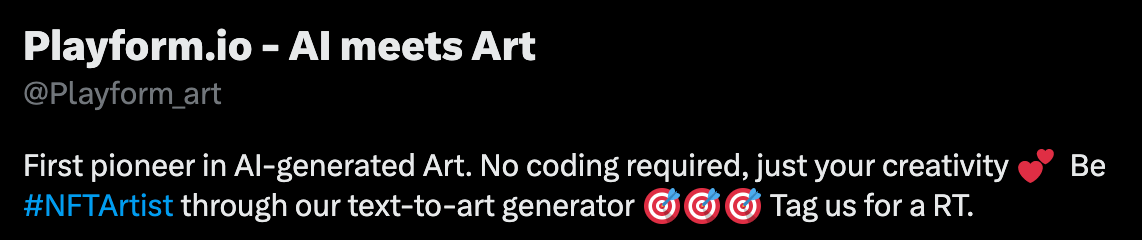}
    \includegraphics[width=\textwidth]{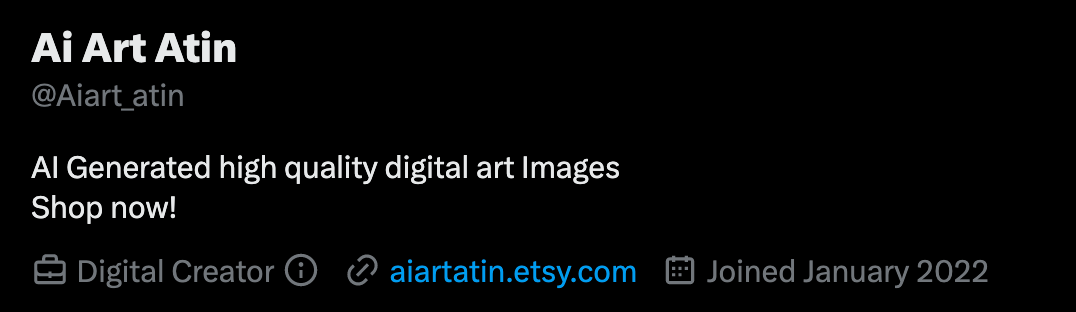}
    \caption{Examples of online references to AI-generated imagery as art}
    \label{fig:twitter}
\end{subfigure}
\hfill
\begin{subfigure}[b]{0.3\textwidth}
    \centering
    \includegraphics[width=\textwidth]{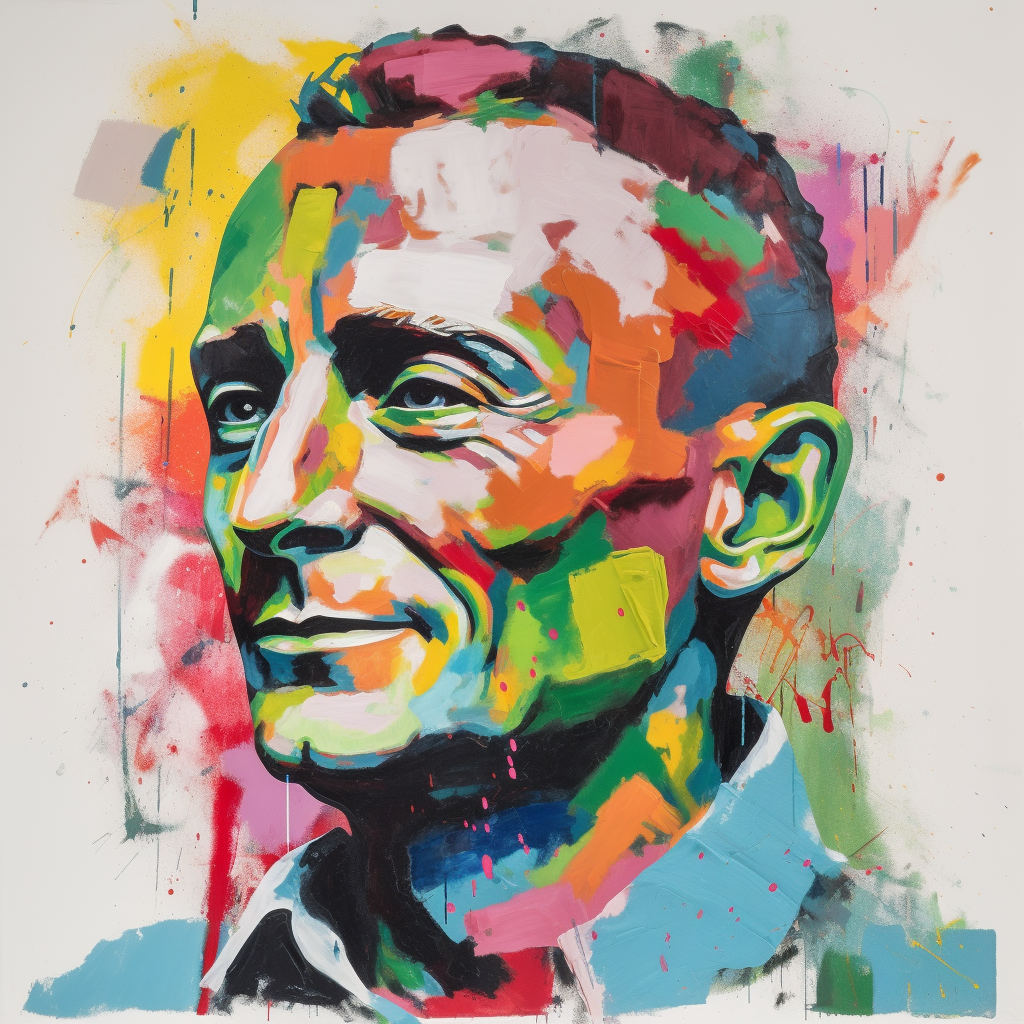}
    \caption{Image generated by Midjourney when prompted with the text: `A work of art'}
    \label{fig:art0}
\end{subfigure}
\hfill
\begin{subfigure}[b]{0.3\textwidth}
    \centering
    \includegraphics[width=0.82\textwidth]{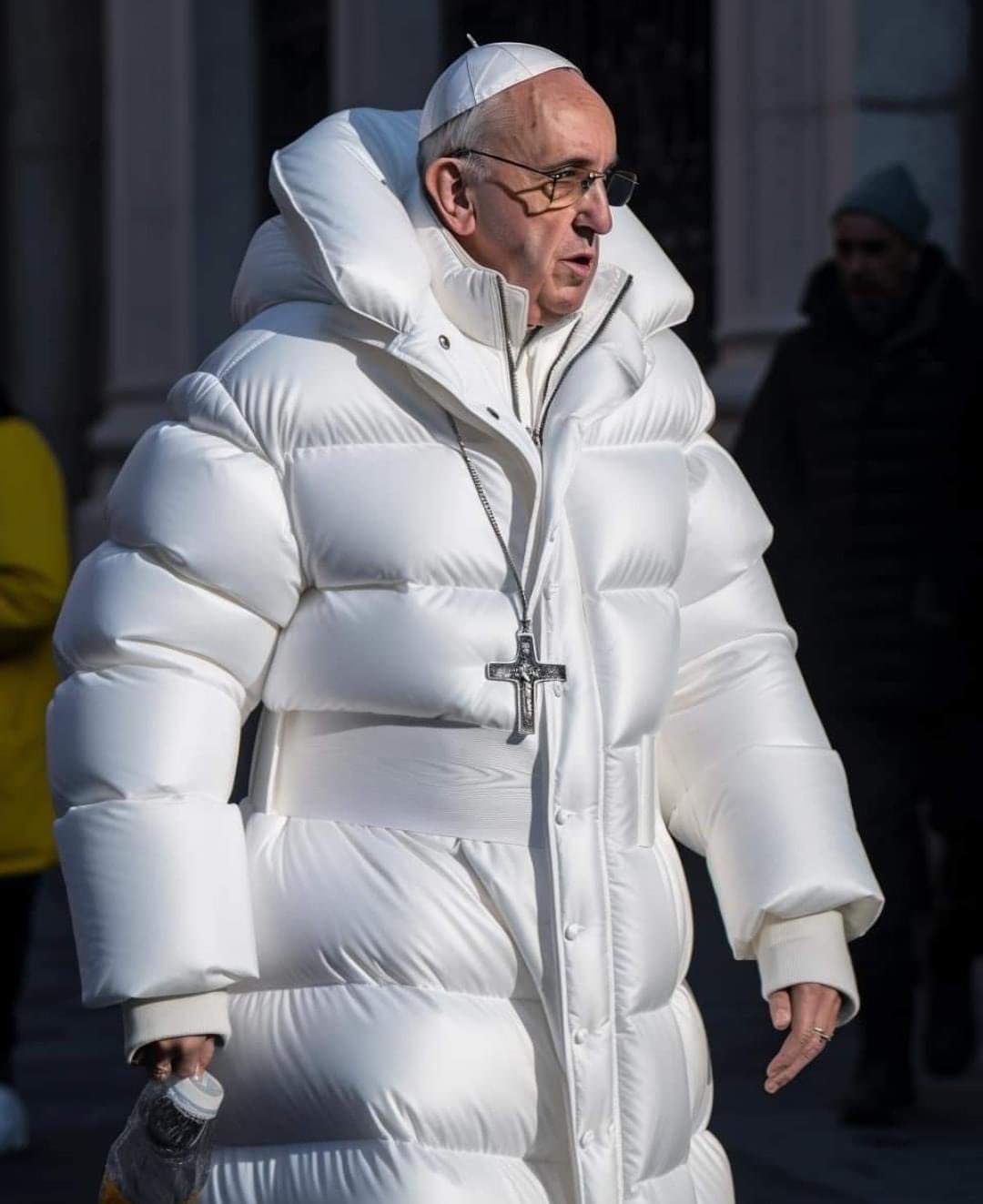}
    \caption{Image generated by Midjourney V5 by Pablo Xavier}
    \label{fig:pope}
\end{subfigure}
\caption{}
\end{figure*}

Within many popular online spaces such as Twitter and Instagram, the recent abundance of artistic imagery produced by text-to-image generative deep learning models has sparked debate as to the nature of these images in the context of more traditional modes of understanding art, and the art world \footnote{We use the term `art' here to primarily refer to visual art.}. We seek to bring some clarity to this discussion using established theoretical frameworks. Text-to-image synthesis requires natural language descriptions that are used to generate corresponding images; the natural language description often being referred to as the `prompt'. Online platforms, such as Midjourney, allow users to generate artistic imagery within seconds, with as little as a single word providing the driving concept for the image synthesis process. The term `art' readily eludes any concrete definition, yet it seems valid given the current climate around the affordances of generative AI, to examine how such a large corpus of digital images has come to be regularly referred to under this title (see figure \ref{fig:twitter}). After over a decade of computational creativity research attempting to encourage a shift in the public perception of computer-generated art, the recent and drastic change in this domain leaves the question open of how and why this has happened now.
The notion of `framing' has been explored within the field of computational creativity with an aim to deepen our understanding of the relationship between the perception of the observer and the artistic artifact being perceived \cite{cook2019framing}. Traditional ideas of framing do not tend to engage with the conceptual baggage of the observer and instead seek to expose aspects of a generative system in the hope that this transparency positively affects the perception of the system, and its output, as creative. We propose a distinct view (but to the same ends) that it is the \emph{viewer} framing, rather than the \emph{artist} framing, that is important to consider in this context. 

\section{`AI Art' - A Philosophy}
There are many frameworks, from the fields of art theory and philosophy, that can help to interrogate the perceptual status of AI-generated imagery in the context of categorising imagery as belonging to the class of `art'. We propose that philosophical theories around the formation of meaning, specifically Wittgenstein's theory of meaning, can inform our understanding of the perception of AI-generated imagery in this contemporary context. 

\subsection{Wittgenstein's theory of meaning}

In his work: `Philosophical Investigations' the German philosopher Ludwig Wittgenstein tackles how it is that, despite the endlessly variable manifestations of things in the world, we can still come to understand similarities between them and can identify groups of occurrences around us \cite{wittgenstein1953philosophical}. He also addresses how it may be that we come to attach meaning to these groups through language, specifically words.

\subsubsection{Wittgenstein's Thought experiment}
To demonstrate this theory, Wittgenstein uses a thought experiment that questions what it means for something to be considered a `game'. The many differences between games clearly do not define whether or not they can still be perceived as qualifying as such, as when trying to describe what the word `game' means, any single defining characteristic of a game can also be found outside of what we perceive `game' to mean. For example, Chess and Solitaire are both considered to be games, even when one involves a winner and a loser and the other does not. So the notion of `game' is a contested concept, as an attempt at any ostensive definition is rejected. Wittgenstein theorises that it is only through a complex web of learnt resemblances and associations that we come to understand a game as belonging to the class of `game'; and not through any concrete definition present in the word itself. Wittgenstein proposes that it is the learnt combination and exemption of traits that contribute to our notion of what the word `game' means, despite there being many ways that people come to understand the umbrella term `games' \cite{wittgenstein1953philosophical}.   

\subsubsection{Family Resemblance}
Wittgenstein uses the analogy of `family resemblance' to illustrate how it is that we identify members of cognitive groups as belonging, whilst discriminating against others. Wittgenstein proposes that we identify one activity as belonging to the category of `game', but not another, in the same way that we might recognise a person as belonging to one family and not another. This is through identifying the resemblances that a person would have with their family due to the fact that they are relations, and the lack of this resemblance otherwise. Factors such as eye colour, surname, hair colour, height, or accent cannot individually be responsible for the decision that someone may be related to someone else, as it then could be concluded that everybody was related. A combination of factors, however, becomes `resemblances' \cite{wittgenstein1953philosophical}. 

\subsubsection{Socio-cultural Context}
Wittgenstein argues that our socio-cultural backdrop is imperative to our understanding of language. He famously theorised that:``If a lion could talk, we could not understand him" \cite{wittgenstein1953philosophical}. A lion is playfully used here, as lions exist within a very different set of social contexts to people. Within Wittgenstein's framework, which anchors meaning in lived experience, it seems consistent to assume that even if a lion could talk, we would not be able to understand what it would say. This is because the language would be derived from a set of socio-cultural circumstances that we are unfamiliar with, not being lions ourselves. Wittgenstein makes the case that it is the matrix of resemblance and associations within a social and cultural context that gives language meaning \cite{wittgenstein1953philosophical}.

\subsection{Art $\Leftrightarrow$  Game}
We propose that the word `game' used in Wittgenstein's thought experiment is interchangeable with the word `art', as this is a similarly contested concept \cite{jordanous2016modelling}. As such, we propose that we come to understand an image as art in the same way that we come to understand an activity as a game - in reference to a plexus of resemblances and associations with existing art.

One could try and make the argument that art is pigment on canvas, but there are examples of sculpture that contradict this, and so on. So, we rely on a self-reflective consciousness of social context, resemblances, and associations in order to recognize something as belonging to a perceptual category such as Art. Just in the same way that all games contribute to the meaning of the word `game', all art contributes to what is meant by the term `art'.

Images and objects that are readily considered to be art manifest in many different forms and so defy any ostensible definition. Given this, we propose that in the same way that Wittgenstein's theory of meaning allows for an explanation for how it is possible that we can identify previously unseen examples of games as belonging to the category of `game', this theory of meaning also helps to explain how it is that AI-generated imagery can be categorised as `art'. It is through our understanding of other art, despite its relentless internal physical differences, that we come to be able to classify something else as art. So it follows, that if an image generated by AI has the abstracted properties and resemblances needed to identify it with this perceptual class, then it becomes a member of that class (Art).

\subsection{Saussure's `Signifier' and `Signified'}
Semiotics is concerned with the making of meaning through the creation and interpretation of `signs'. Signs can be words, images, sounds, acts, or objects. According to the American  philosopher, mathematician, and scientist Charles Sanders Pierce: ``We think only in signs'' \cite{peirce1992writings}. 

The Swiss Linguist Ferdinand de Saussure and, later on, the writings of the French literary theorist Roland Barthes (who famously declared `The Death of the Author') influenced thought on the role of semiotics in the deconstruction and extraction of meaning in the social context. This is through the recognition of a dyadic system of signs.  For Saussure, semiotics breaks down the factors in our external environment into signs, which are comprised of a `signifier' and a `signified'. The `signifier' is the physical form the sign takes, and the `signified' is the concept that it represents. The sign is the result of the association of the signifier with the signified, and the relationship between them is referred to as `signification' \cite{saussure:course}. The meaning of signs is said not to be inherent \emph{in} the words, images, sounds, acts or objects themselves, but rather in the meaning we attach to them. Anything can be a sign as long as someone interprets it as referring to something else. We often interpret signs unconsciously, relating them to familiar systems of conventions: ``A sign is the basic unit of language (a given language at a given time). Every language is a complete system of signs" \cite{saussure:course}.

\subsection{Image as `signifier', Art as `signified'}
The question of how a digital image can be removed from its status as simply an `image' and be interpreted within a new field of understanding where it is perceived to have a meaning beyond its basic function, we propose, starts in the interaction between the `consumer' of the image, and the image itself. This echoes the ideas explored in work by Colton and Wiggins \cite{colton2012computational}, where Computational Creativity is centered around observers and audiences. The way in which an image is perceived is similarly relevant to examine here, as it is at this point of consumption by an audience that an understanding can be formed by the observer, and where a decision as to the status of an image can be decided (i.e. an image is given the status of `art').
According to the dyadic theory of semiotics, images have the potential to become art if they are a signifier for the concept of art. In this case, the image is the `signifier' and the idea of art is the signified. The signifier is able to function as such as it alludes to the matrix of associated properties (Wittgenstein) that is signified. Figure \ref{fig:art0} shows an example of this idea. We suggest that if meaning arises in the consumption of a dyadic system of signs, and art consists of these very factors, then it follows that an image can be understood as art through a participation in this system and the reciprocal nature of the interaction between the perceiver and the sign being perceived. 
Within the framework of Wittgenstein's theory, it is possible to understand how an image may come to be understood as art given the social context. Through the linguistic phenomena of semiotics, it is possible to understand the relationship between an image as a signifier, and the concept anchored to it as the signified - which is shaped by the given social context. Given this, it seems that it is necessary for there to be a much wider presence in a work of art, of a space for the intuitive understanding of the perceiver, but also a reference to something `learnt'. We propose that it is the \emph{reading} of an image that facilitates its attribution to the class of art.


Figure \ref{fig:art0} is an example of an AI-generated image (made using Midjourney) that demonstrates the many different visual properties in an image that AI may generate in order to be a signifier to the concept of art and satisfy its internal categorisation system given a prompt (i.e. the user asks for `a work of art', as shown in figure \ref{fig:art0}). A brief visual analysis exposes the following properties associated with art genres: portrait painting (a 3/4 profile of a man is shown), realism (the portrait features are mildly abstracted), graffiti art (colourful and chaotic spray paint marks and splatters), Impressionism (colour blocks in the face give the impression of form, and create highlights and shadow - but are not true to a real-life depiction), Pop Art (bold, bright, multi-coloured areas in direct contrast with areas of black). A closer analysis reveals that the colours in the image correspond to colour theory. The image is mostly constructed from two sets of complementary colours: blue/orange and red/green. Furthermore, different shades of these colours are used to create highlights and shadows. These aesthetic properties are indicators within the signifier (the image) that signify these well know conceptions of visual art.


Figure \ref{fig:pope} shows an example of an image generated by Midjourney that doesn't explicitly hook into the visual properties associated with artworks, and yet has cultural significance within a social context. In fact, when the image of the Pope first became viral on Twitter, many people believed it to be a real photograph \cite{buzzfeednews}. It only surfaced later on that it was in fact an AI-generated image. This phenomenon is an example of the importance of the audience's perception of an image regarding the classification of that image within the discourse around it. 




\subsection{Duchamp's `Readymades'}

Marcel Duchamp's concept of the `readymade' was a significant development in the history of modern art. The term refers to everyday mass-produced objects that were selected and presented as works of art. The act of choosing these objects, and presenting them in the context of an art exhibition, was challenging traditional notions of what could be considered art and questioning the value of artistic skill and craftsmanship.
The chosen objects were often prefabricated and were separated from their intended `mundane' use by being brought into the art gallery context and discourse \cite{duchamp:readymade}.
We propose that AI-generated imagery can be a conceptual extension of the phenomena of the `readymade', by virtue of them being mass-produced and can challenge our existing notions of what can constitute an art artifact. As we will explore in the next section, many of 
Duchamp's readymades were intended to challenge traditional perceptions of the role of craftsmanship in art. Mass-produced AI-generated imagery poses a similar challenge to the kinds of skills needed to create culturally significant and artistic imagery. 


\subsubsection{Mass production and the `Bach Faucet'}
Text-to-image models are systems that are capable of synthesising potentially unlimited amounts of new and high-quality digital imagery, which we can view as a kind of `Bach faucet', a term coined by Kate Compton to refer to a situation in which: `a generative system produces an infinite amount of content that is of equal or better quality than a culturally significant original...since the endless supply of this content makes it no longer rare, it decreases its value'\cite{compton:bachfaucet}. This phenomenon represents the inverse of the value transaction inherent in the process of creating readymades. In the case of the readymade, an object of low scarcity and value is transmuted into a scarce object with high value. AI however is able to take scarce and high-value artifacts (artworks) and mass produce images of equal or increased quality which lowers scarcity, decreasing the overall value. 

\subsection{Latent space imbues generated imagery with artistic signifiers}
It is timely to intersect this theory with notions of machine learning and how complexity is abstracted during the training, and image synthesis, process.

We propose that image synthesis algorithms, that are trained on large-scale data sets of corresponding signs (or patterns \cite{alpaydin2010introduction}), are able to abstract the complexity of the matrix of multiple associated properties (or signifiers), that according to Wittgenstein and Saussure accumulates as an understanding of what art comes to mean, into a `latent space'. Any image generated from this space, that relates to `art' (via a text prompt), is imbued with the properties learnt from the signs in the training data which act as signifiers to the concept of art. We propose that, as a result, when we encounter these properties in AI-generated imagery, we are easily able to associate the image (signifier) with the web of resemblances and associations that we bring to the interaction regarding art (the signified). With some degree of inevitability, given the potential for exposure to the images in the data (pre-training), the generated images match our perception of `art' - as the model is trained on many thousands of existing examples of art. This is why, we propose, that so many AI-generated images are considered to belong to the category of art.  

\subsection{Kate Compton's `Liquid Art'}
The term `Liquid Art' was disseminated at an invited talk given by Kate Compton at MIT in the autumn of 2022: ``The ARTIFACT is dead. All that remains is the EXPERIENCE. Welcome to the world of liquid art'' \cite{aialchemy}. This term describes a new form of art experience, specifically in online spaces, in the wake of the invention of text-to-image generative AI systems \cite{aialchemy}. 


Liquid art is described as a space of potential art artefacts, that is moved through by surfing and filters, and that is experienced in streams or overwhelming waves (most commonly in an online space). Liquid art is the phenomenon of being exposed to mass-produced artefacts, such as AI-generated imagery, in a space where its abundance means that the experience of `surfing' this media becomes the experience of art. In this framework, art is a verb - as art becomes the experience of moving through this possibility space \cite{aialchemy}. This is opposed to more traditional forms of `Solid' art, where there is an art artifact of fixed form.
Liquid art has implications for images and their function as signs in our language system, including visual language systems, as the abundance of imagery and the properties of this imagery (high-quality artistic imagery) shifts the social context and environment for the sign within art, and the context within which signifiers of art function.  



\subsubsection{Liquid Art and the Readymade}
We propose that, out of this endlessly generated sea of imagery, an image can be selected based on the ways in which it functions as a sign. Another way to conceptualise this would be to class the output space of the image generator as a mass-produced artefact (a space of images) and the act of prompting and identifying an image as `good' is what creates the readymade.
We propose that because the history of a generated image (the transmutation of meaning from the training data into the latent space, re-manifested as an image in a liquid perceptual possibility space, from which it is then selected as a readymade for its quality as a sign within visual art) is so rich, that there is an increase in value in the selection of generated imagery in particular as a new era of readymade, because traditional readymades as a concept (and also an aesthetic to be signified) are already a part of the web of understanding that resonates with this selection process as they are an established part of art history now. At the same time, these images can still serve the function of challenging traditional notions of what can be considered as art (as with \ref{fig:pope}) and question the value of artistic skill and craftsmanship as the traditional readymades did. 

\section{Conclusion and Future Work}
AI-generated imagery can come to be perceived as art according to its perception as a sign by an observer within their social context. The use of mass-produced AI-generated imagery as art can be seen as a conceptual extension of the `readymade' within in a complex contemporary context, where the selection and presentation of the image can challenge even current notions of what is required for the experience of `art'. We suggest that future work could go on to discuss memes as readymades, based on the way that they involve the use of mass-produced imagery and hold relevance to the socio-cultural backdrop of the time. 

\bibliographystyle{iccc}
\bibliography{iccc}

\begin{thebibliography}{}

\bibitem[\protect\citeauthoryear{Alpaydin}{2010}]{alpaydin2010introduction}
Alpaydin, E.
\newblock 2010.
\newblock Introduction to machine learning.
\newblock {\em The Massachusetts Institutes of Technology Press}.

\bibitem[\protect\citeauthoryear{Colton and
  Wiggins}{2012}]{colton2012computational}
Colton, S., and Wiggins, G.~A.
\newblock 2012.
\newblock Computational creativity: The final frontier?
\newblock In {\em Proceedings of Ecai}.

\bibitem[\protect\citeauthoryear{{Compton, K}}{2022}]{aialchemy}
{Compton, K}.
\newblock 2022.
\newblock Terrible together: A creativity manifesto.
\newblock http://aialchemy.media.mit.edu/terrible-together.html.
\newblock Accessed: April 17, 2023.

\bibitem[\protect\citeauthoryear{Compton}{2013}]{compton:bachfaucet}
Compton, K.
\newblock 2013.
\newblock The bach-pedal-point faucet: A computational model of musical
  harmony.
\newblock In {\em Proceedings of the International Conference on Computational
  Creativity}.

\bibitem[\protect\citeauthoryear{Cook \bgroup et al.\egroup
  }{2019}]{cook2019framing}
Cook, M.; Colton, S.; Pease, A.; and Llano, M.~T.
\newblock 2019.
\newblock Framing in computational creativity-a survey and taxonomy.
\newblock In {\em Proceedings of the International Conference on Computational
  Creativity}.

\bibitem[\protect\citeauthoryear{Duchamp}{1957}]{duchamp:readymade}
Duchamp, M.
\newblock 1957.
\newblock The creative act.
\newblock {\em Art News} 56(4).

\bibitem[\protect\citeauthoryear{Jordanous and
  Keller}{2016}]{jordanous2016modelling}
Jordanous, A., and Keller, B.
\newblock 2016.
\newblock Modelling creativity: Identifying key components through a
  corpus-based approach.
\newblock {\em PloS one} 11.

\bibitem[\protect\citeauthoryear{Peirce}{1992}]{peirce1992writings}
Peirce, C.~S.
\newblock 1992.
\newblock {\em Writings of Charles S. Peirce: A Chronological Edition}.
\newblock Indiana University Press.

\bibitem[\protect\citeauthoryear{Saussure}{1983}]{saussure:course}
Saussure, F.~d.
\newblock 1983.
\newblock {\em Course in General Linguistics}.
\newblock London : G. Duckworth.

\bibitem[\protect\citeauthoryear{Stokel-Walker}{2023}]{buzzfeednews}
Stokel-Walker, C.
\newblock 2023.
\newblock This ai program was tasked with creating images.
\newblock Accessed: April 17, 2023.

\bibitem[\protect\citeauthoryear{Wittgenstein}{2010}]{wittgenstein1953philosophical}
Wittgenstein, L.
\newblock 2010.
\newblock {\em Philosophical investigations}.
\newblock John Wiley \& Sons.

\end{thebibliography}

\end{document}